\documentclass[10pt,twocolumn,letterpaper]{article}

\usepackage{multirow}
\usepackage{bbm, dsfont}
\usepackage[pagenumbers]{cvpr} 

\newcommand{\set}[1]{\mathcal{#1}}

\newcommand\blfootnote[1]{%
  \begingroup
  \renewcommand\thefootnote{}\footnote{#1}%
  \addtocounter{footnote}{-1}%
  \endgroup
}
\usepackage[dvipsnames]{xcolor}

\definecolor{cvprblue}{rgb}{0.21,0.49,0.74}
\usepackage[pagebackref,breaklinks,colorlinks,citecolor=cvprblue]{hyperref}

\def\paperID{6903} 
\def\confName{CVPR}
\def\confYear{2024}

\title{Improving the Generalization of Segmentation Foundation Model under Distribution Shift via Weakly Supervised Adaptation}

\author{
Haojie Zhang$^1$\footnotemark
\and
Yongyi Su$^{1}$\footnotemark[1]
\and
Xun~Xu$^2$\footnotemark
\and
Kui Jia$^3$
\and
$^1$South China University of Technology \\
$^2$Institute for Infocomm Research, A*STAR \\
$^3$School of Data Science, The Chinese University of Hong Kong, Shenzhen \\
{\tt\small{\url{https://github.com/Zhang-Haojie/WeSAM}}}
}

\begin{document}

\maketitle

\begin{abstract}
\vspace{-0.1cm}

The success of large language models has inspired the computer vision community to explore image segmentation foundation model that is able to zero/few-shot generalize through prompt engineering. Segment-Anything~(SAM), among others, is the state-of-the-art image segmentation foundation model demonstrating strong zero/few-shot generalization. Despite the success, recent studies reveal the weakness of SAM under strong distribution shift. In particular, SAM performs awkwardly on corrupted natural images, camouflaged images, medical images, etc. Motivated by the observations, we aim to develop a self-training based strategy to adapt SAM to target distribution. Given the unique challenges of large source dataset, high computation cost and incorrect pseudo label, we propose a weakly supervised self-training architecture with anchor regularization and low-rank finetuning to improve the robustness and computation efficiency of adaptation. We validate the effectiveness on 5 types of downstream segmentation tasks including natural clean/corrupted images, medical images, camouflaged images and robotic images. Our proposed method is task-agnostic in nature and outperforms pre-trained SAM and state-of-the-art domain adaptation methods on almost all downstream tasks with the same testing prompt inputs.

\blfootnote{\textsuperscript{*} Equal contribution.}
\blfootnote{\textsuperscript{\dag} Correspondence to Xun~Xu: alex.xun.xu@gmail.com.}

\end{abstract}

\begin{figure}[t]
    \centering
\includegraphics[width=1.01\linewidth]{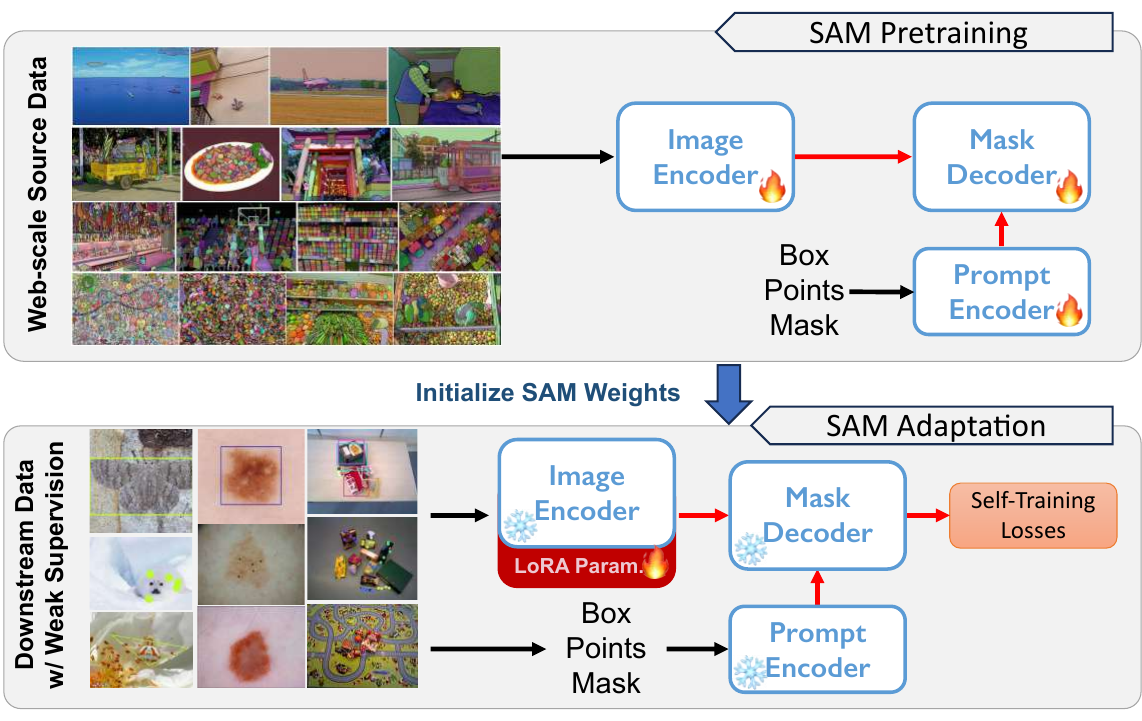}
    \caption{Segment Anything Model was pre-trained on a large-scale dataset but exhibits awkward performance on diverse downstream segmentation tasks. We adapt SAM through weak supervision to enhance its generalization capabilities.\vspace{-0.3cm}}
    \vspace{-0.4cm}
    \label{fig:enter-label}
\end{figure}

\vspace{-0.7cm}

\section{Introduction}

The success of large language models is partially attributed to the strong capability of zero-shot generalization.
This has called for numerous attention into developing foundation model for computer vision tasks where one of the key concern also lies in the generalization ability~\cite{kim2021domain,oquab2023dinov2,radford2021learning}. In particular, as a promptable image segmentation foundation model with strong zero-shot generalization, the Segment-Anything model~(SAM)~\cite{kirillov2023segment} was developed by training on billions of annotated masks. 
Despite the overwhelming size of dataset used for training, SAM was found to behave awkwardly on certain out-of-distribution downstream tasks, including camouflaged segmentation, medical segmentation, adversarial attacks, visual corruptions, etc.~\cite{chen2023sam, huang2023robustness}. This is likely caused by the distribution shift between the training data, i.e. the SA-1B dataset~\cite{kirillov2023segment}, and the challenging testing datasets. These observations motivate us to explore tangible solutions to make SAM more robust against real-world and more diverse downstream tasks.

Traditional paradigm towards improving model robustness and generalization often involves costly re-training. Methods are frequently customized to combat specific domain shifts. For example, domain randomization was developed to improve generalization across real testing domains~\cite{chen2021understanding}, while adversarial training is commonly used to enhance robustness against attacks~\cite{szegedy2013intriguing}.
Applying these techniques to large foundation model is nonetheless impractical due to the enormous computing resources required for training the model on the web-scale training data. Therefore, instead of re-training the model, we opt for a more computation friendly paradigm by adapting or finetuning a pre-trained foundation model to downstream dataset.

We specify three major challenges when adapting pre-trained foundation model to a new data distribution. First, traditional unsupervised domain adaptation~\cite{ganin2015unsupervised} paradigm requires access to both source and target datasets which may not be feasible due to privacy issue and computation cost~\cite{liang2020we}. Second, updating all model weights for adaptation is often superior in performance, however, is constrained by the prohibitive memory cost due to the large model size~\cite{hu2022lora}.
Finally, unsupervised adaptation could be still very challenging due to the absence of label information on the target domain~\cite{tsai2018learning,zhang2019category,zou2019confidence}. 

To tackle the above challenges, we propose a self-training based adaptation approach to synergize the exploitation of weak supervision and unlabeled data on target domain. 
Specifically, we first alleviate the dependence on source domain data by adopting a self-training based source-free domain adaptation strategy. The self-training approach first make segmentation predictions, a.k.a. pseudo labels, on the target domain data. The pseudo labels are  used for supervising the update of segmentation model. 
As self-training is fragile due to incorrect pseudo labels, a.k.a. confirmation bias~\cite{arazo2020pseudo}, we introduce a frozen source model as the anchor network to regularize the update of segmentation model.
To further mitigate the high computation cost of updating the full model weights, we apply a low-rank weight decomposition to encoder weights and the backpropagation flows through the low-rank shortcut only. Finally, to further improve the effectiveness of source-free domain adaptation, weak supervisions, e.g. sparse point-wise annotation, on the target domain are introduced to provide stronger cues for domain adaptation. We reveal that the weak supervisions are naturally compatible with the prompt encoder within SAM. With the weak supervisions as prompt, we are endowed with more localized and less ambiguous pseudo predictions for self-training. The adapted model has demonstrated much stronger generalization capability on multiple downstream tasks.

We summarize the contributions of this work as follows.

\begin{itemize}
    \item We are motivated by the  generalization issue of segment-anything~(SAM) model to diverse downstream segmentation tasks and propose a task-agnostic solution to adapt SAM through self-training with no access to source dataset.
    \item We exploit weak supervisions, including bounding box, point-wise annotation and coarse segmentation masks, to improve the effectiveness of adaptation. The weak supervision are fully compatible with the prompt encoder of SAM.
    \item Extensive experiments on 5 types of downstream instance segmentation tasks demonstrate the effectiveness of the proposed weakly supervised adaptation approach.
\end{itemize}

\begin{figure*}[t]
    \centering
\includegraphics[width=0.91\linewidth]{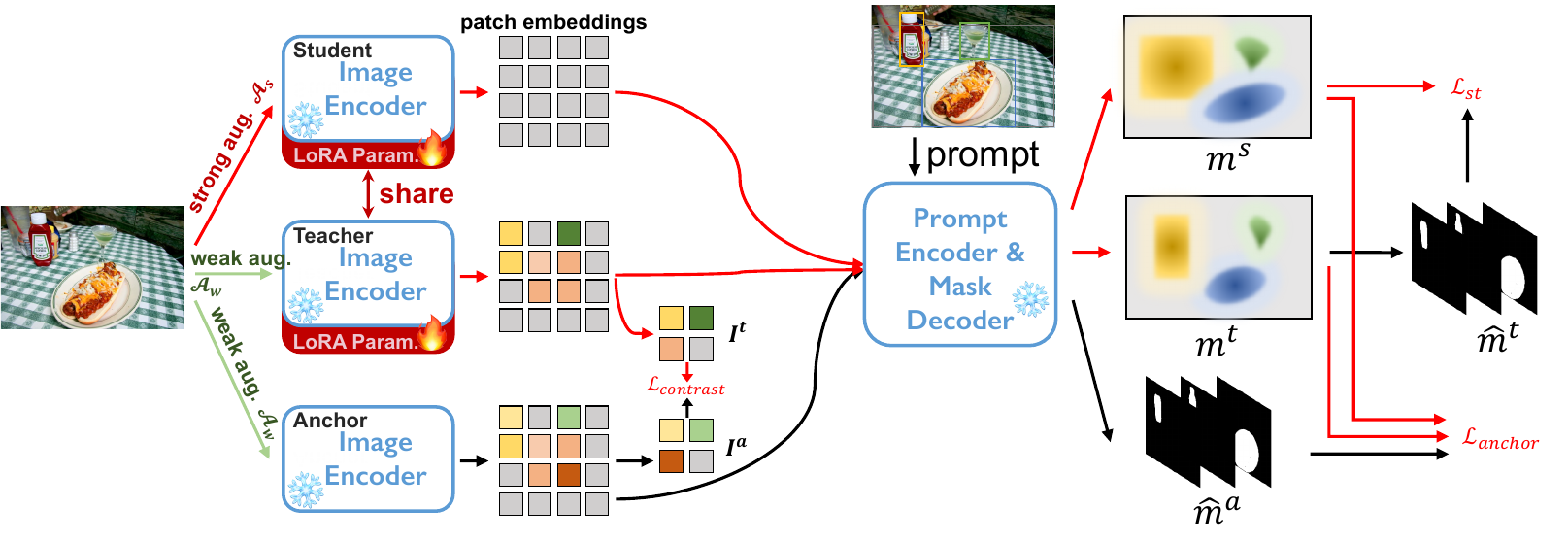}
    \vspace{-0.3cm}
    \caption{The proposed self-training architecture with anchor network regularization and contrastive loss regularization. Red arrows indicates the backpropagation flow.}
    \vspace{-0.5cm}
    \label{fig:self-training}
\end{figure*}

\vspace{-0.2cm}
\section{Related Work}
\vspace{-0.2cm}

\noindent\textbf{Image Segmentation Foundation Model}: 
The success of deep learning is attributed to the increasing size and neural networks and huge amount of training data. Recent successful computer vision models, e.g. image segmentation and object detection, follow a practice of finetuning from encoder network pre-trained on large image dataset, e.g. ImageNet~\cite{deng2009imagenet}. Despite achieving impressive results on standard benchmark datasets, e.g. Psacal VOC~\cite{everingham2015pascal} and COCO~\cite{lin2014microsoft}, the zero-shot and few-shot generalization ability is often limited. Inspired by the revolutionizing language/vision-language foundation model pre-trained on web-scale datasets~\cite{brown2020language,radford2021learning,zhang2024dual}, a new opportunity awaits for developing a generalizable vision foundation model. This motivates the emergence of vision foundation model pre-trained on huge dataset. SAM~\cite{kirillov2023segment} and DINO~v2~\cite{oquab2023dinov2}, to name a few. Among these foundation models, SAM stands out for its ability to enable zero-shot segmentation using prompt engineering. Therefore, we pay particular attention to SAM in this work.
Since the inception of SAM, numerous attempts are made to validate the robustness of SAM under more challenging scenarios, such as medical images \cite{ma2023segment} \cite{zhang2023customized}, camouflaged objects \cite{tang2023can}, glass (transparent objects and mirrors) \cite{han2023segment} and pose estimation~\cite{lin2023sam6d}. Despite these early attempts to discover the weakness, there are very few matured solutions~\cite{chen2023sam} to improve the generalization of SAM to challenging downstream tasks. In this work, we aim to fill this gap by proposing principled solutions to enhance the robustness of SAM against downstream tasks subject to significant distribution shift.

\noindent\textbf{Source-Free Domain Adaptation}: 
Unsupervised Domain Adaptation (UDA)\cite{ganin2015unsupervised, long2015learning, jing2022adversarial, tzeng2017adversarial, zhang2020unsupervised} has emerged as a means to address the domain shift between source training data and target testing data. This enables models to be trained on cost-effective and easily annotated data, facilitating the transfer to more valuable but challenging-to-label datasets. However, UDA methods face limitations when source domain data is inaccessible due to privacy concerns or storage constraints. In response to this, Source-Free Domain Adaptation (SFDA)\cite{liang2020we, li2020model, kundu2020universal, yang2021generalized, xia2021adaptive, liu2021ttt++} has been proposed to reduce the reliance on source data in domain adaptation (DA) methods. 
SHOT~\cite{liang2020we} represents an early effort to alleviate domain shift without access to source data, achieved through self-training on unlabeled target data. 
Numerous prior works~\cite{kim2021domain, ding2023proxymix, ahmed2022cleaning, qu2022bmd, shen2022benefits, chu2022denoised, liang2021source} have explored the effectiveness of a teacher-student like self-training architecture in domain shift mitigation. The similar self-training framework \cite{huang2023weakly} simultaneously trains on multiple sets of augmented data, utilizing teacher-student cycle consistency for mutual learning of the two networks.
Other methods generate source-like samples~\cite{kurmi2021domain, qiu2021source, liu2021source}, construct intermediate domains~\cite{xia2021adaptive, yang2023casting}, or perturb target domains~\cite{xiong2021source, zhang2021source, jing2022variational, li2021divergence} to enhance model generalization. Recent approaches~\cite{qiu2021source, wang2022cross} leverage contrast loss between target samples and source domain prototypes. Recent attempts to source-free domain adaptation for image segmentation aims to minimize feature discrepancy between source and target domains~\cite{liu2021source,bateson2022source,kundu2021generalize}.
In this work, we aim to adapt SAM to downstream tasks without accessing to the source domain data to avoid the high computation overhead and potential privacy issues for foundation models to be released in the future.

\noindent\textbf{Weakly Supervised Domain Adaptive Segmentation}: 
Unsupervised domain adaptation often follows a self-training paradigm~\cite{liang2020we}, which is limited by the quality of pseudo labels. Instead of relying on the pseudo predictions, weakly supervised domain adaptive semantic segmentation~(WDASS) exploits limited weak supervisions in the form of bounding box, points or coarse segmentation masks~\cite{paul2020domain,wang2019weakly, das2023urban,das2023weakly}.  Among these works, \cite{paul2020domain} employs both image and point labels, suggesting an adversarial alignment of features at the pixel level to address WDASS. In contrast, \cite{wang2019weakly} relies on bounding boxes as weak labels and employs adversarial learning to achieve domain-invariant features. Additionally, \cite{das2023urban} utilizes self-training and a boundary loss to enhance performance in WDASS with coarse labels. More recently, \cite{das2023weakly} utilizes weak label for
aligning the source-target features and outperforms previous methods to achieve competitive performance compared
to supervised learning. We observe that the common weak supervisions are inherently compatible with the SAM model as prompt input. Therefore, we propose to adapt SAM with weak supervisions in a seamless way.

\section{Methodology}
\vspace{-0.2cm}

In this section, we first provide an overview of segment-anything model. Then, we introduce the self-training based adaptation framework. We elaborate how weak supervision could help establish correspondence between pseudo predictions for effective self-training. Lastly, we introduce the low memory footprint way of updating model weights.

\vspace{-0.2cm}
\subsection{Overview of Segment Anything Model}
\vspace{-0.2cm}

The segment-anything model consists of three major components, the encoder network $z=f(x;\Theta)$, prompt encoder $e=g(p;\Omega)$ and mask decoder $h(z,e;\Phi)$. The image encoder is pre-trained using masked autoencoder~(MAE)~\cite{he2022masked}. The whole SAM model is further fine-tuned on the web-scale labeled training set, SA-1B~\cite{kirillov2023segment}, with 1.1B labeled masks. A combination of focal loss and dice loss are used for training SAM. During inference, the testing image $x$ is first encoded by the image encoder $z=f(x;\Theta)$. Given encoded prompts $e$, a light-weight mask decoder makes three levels of segmentation predictions. In this work, we are motivated by the challenge of deploying SAM on many downstream tasks and propose to adapt SAM to downstream segmentation tasks with weak supervision without requiring access to source domain training data.

\vspace{-0.2cm}
\subsection{Source-Free Adaptation as Self-Training}
\vspace{-0.2cm}

Provided an unlabeled target domain dataset $\set{D}_{T}=\{x_i\}$ and the pre-trained encoder network $f(x;\Theta)$, we adopt a teacher-student architecture for self-training. As illustrated in Fig.~\ref{fig:self-training}, we maintain three encoder networks, namely the anchor network $f(x;\Theta^a)$, the student network $f(x;\Theta^s)$ and the teacher network $f(x;\Theta^t)$, where the student and teacher networks share the weights $\Theta^s=\Theta^t$. For each sample $x_i$, we apply one random weak data augmentation $\mathcal{A}_w(x_i)$ as the input to anchor and teacher networks, and one random strong data augmentation $\mathcal{A}_s(x_i)$ as the input to student network. 
Through the three encoder networks, three feature maps are obtained respectively, as $\mathcal{F}^a=f(\mathcal{A}_w(x_i), \Theta^a)$, $\mathcal{F}^s=f(\mathcal{A}_s(x_i), \Theta^s)$ and $\mathcal{F}^t=f(\mathcal{A}_w(x_i), \Theta^t)$, where $\mathcal{F} \in \mathbbm{R}^{D\times H\times W}$ and $D$ refers to feature dimension. In the decoder network, given a number $N_p$ of prompts, e.g. bounding boxes, points or coarse segmentation masks, a set of foreground masks for the instance segmentation results would be deduced, $\mathcal{Y}^a=\{y^a_j\}_{j=1\cdots N_p}$, $\mathcal{Y}^s=\{y^s_j\}_{j=1\cdots N_p}$, $\mathcal{Y}^t=\{y^t_j\}_{j=1\cdots N_p}$, where each mask is then normalized between 0 and 1 by sigmoid function respectively, i.e. $m_j=Sigmoid(y_j), m_j\in\left[0,1\right]^{H\times W}$. We further denote a binarized segmentation mask with a threshold of 0.5 as $\hat{m}_j=\mathbbm{1}(m_j>0.5), \hat{m}_j \in \{0, 1\}^{H\times W}$. 
Based on these notations, we elaborate three sets of adaptation objectives for self-training purpose.

\noindent\textbf{Teacher-Student Self-Training Loss}:
We first introduce the Self-Training loss to update student/teacher network. Self-Training is widely used in semi-supervised learning~\cite{sohn2020fixmatch} and recently is demonstrated to be very effective for source-free domain adaptation~\cite{liang2021source}. In this work, we continue the idea of self-training in classification task and perform focal loss~\cite{lin2017focal} and dice loss~\cite{milletari2016v} for the supervision of student's outputs with binarized teacher's predictions as follows, where $\gamma$ controls the focus on hard training pixels and $\epsilon$ is a small value to avoid dividing by zero.
\vspace{-0.3cm}
\begin{equation}
    \resizebox{0.9\linewidth}{!}{
    $
        \begin{split}
        \mathcal{L}^{focal}_{st} & = -\sum_{j=1\cdots N_p} \frac{1}{HW} \sum_{h,w} \mathbbm{1}(\hat{m}^t_{jhw}=1) \cdot (1-m^s_{jhw})^\gamma log(m^s_{jhw}) \\
            & + \mathbbm{1}(\hat{m}^t_{jhw}=0) \cdot {m^s_{jhw}}^\gamma log(1-m^s_{jhw}),
        \end{split}
    $
    }
\end{equation}
\vspace{-0.4cm}
\begin{equation}
    \mathcal{L}^{dice}_{st} = \sum_{j=1\cdots N_p} 1 - \frac{2\sum_{h,w} m^s_{jhw} \cdot \hat{m}^t_{jhw} + \epsilon}{\sum_{h,w} m^s_{jhw} + \sum_{h,w} \hat{m}^t_{jhw} + \epsilon},
\end{equation}

\noindent\textbf{Anchored Loss for Robust Regularization}: Training the networks with  self-training loss alone is vulnerable to the accumulation of incorrect pseudo labels predicted by teacher network, a.k.a. confirmation bias~\cite{arazo2020pseudo}. Observations are made that the performance will drop after long iterations of self-training alone~\cite{su2023revisiting}. Existing source-free domain adaptation approaches often adopt additional constraints to prevent the negative impact of self-training, e.g. uniform distribution over predictions~\cite{liang2021source}. Without making such explicit assumptions, we propose to introduce the regularization by an anchor loss which minimizes the dice loss between an anchor network, which has frozen source model weights, and student and teacher networks respectively in Eq.~\ref{eq:anchor}. The frozen anchor network serves as the knowledge inherited from source domain and too much deviation between the source model and self-training updated model is discouraged to prevent the model from collapsing.

\vspace{-0.3cm}
\begin{equation}\label{eq:anchor}
\resizebox{0.9\linewidth}{!}{
$
\mathcal{L}_{anchor} = \lambda^{dice}_{stu}\mathcal{L}^{dice}(m^s, \hat{m}^a) + \lambda^{dice}_{tea}\mathcal{L}^{dice}(m^t, \hat{m}^a).
$
}
\end{equation}
\vspace{-0.3cm}

\begin{figure}[t]
    \centering
\includegraphics[width=1.02\linewidth]{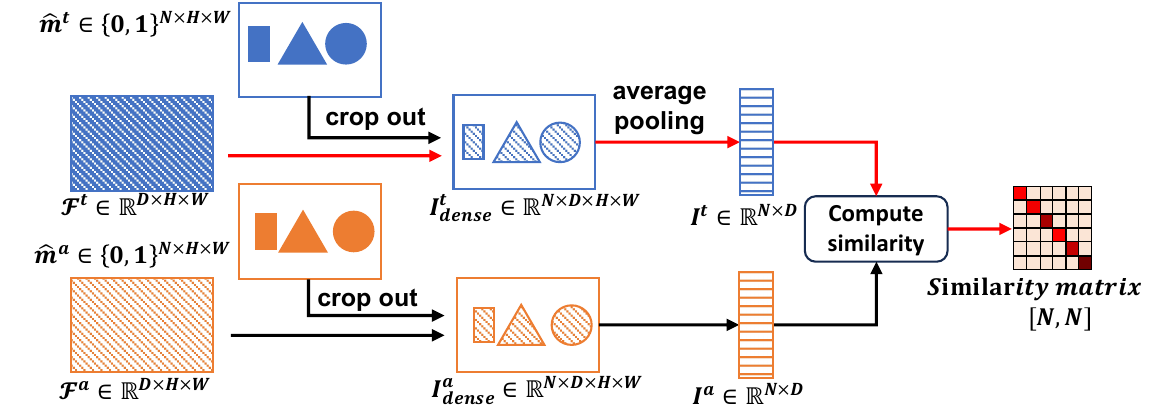}
    \vspace{-0.5cm}
    \caption{Illustration of contrastive loss between two views. \vspace{-0.2cm}}
    \vspace{-0.3cm}
    \label{fig:ContrastiveLoss}
\end{figure}

\noindent\textbf{Contrastive Loss for Encoder Output Regularization}: The above two training objectives are performed in the output space of the decoder. As we empirically revealed in the experiment section, updating encoder network is the most effective way to adapt SAM, it is necessary to apply regularization directly to the feature map as output from the encoder network.
Specifically, as illustrated in Fig.~\ref{fig:ContrastiveLoss}, we first crop out each instance's feature $I_j$ from the feature map based on its prediction mask in anchor and teacher branches.
\vspace{-0.2cm}
\begin{equation}
    I_j=\frac{\sum_{h,w} \overline{\mathcal{F}}_{hw} \cdot \hat{m}_{jhw}}{\sum_{h,w}\hat{m}_{jhw}},
\end{equation}
\vspace{-0.2cm}

where $\overline{\mathcal{F}}$ is the L2 normalized feature map, i.e. $\overline{\mathcal{F}}_{hw}=\frac{\mathcal{F}_{hw}}{|\mathcal{F}_{hw}|}$. We further define the instance features, which are cropped out by the masks used the same prompt in two branches, as positive pairs, and the instances used different prompts as negative pairs to construct the contrast loss. The final contrastive loss is given as below, where $\tau$ is the temperature scaling. 

\vspace{-0.4cm}
\begin{equation}
    \mathcal{L}_{contrast}=-log \frac{\sum_j exp( I^a_j \cdot I^t_j / \tau )}{ \sum_j \sum_{j'\neq j} exp( I^a_j \cdot I^t_{j'} / \tau)}
\end{equation}
\vspace{-0.2cm}

\noindent\textbf{Total Loss}: We combine the above three loss functions as the final source-free domain adaptation loss.
\begin{equation}
    \mathcal{L} = \lambda^{focal}_{st} \mathcal{L}_{st}^{focal} + \mathcal{L}_{st}^{dice} + \mathcal{L}_{anchor} + \mathcal{L}_{contrast}.
\end{equation}

\subsection{Prompt Generation for Self-Training}
\vspace{-0.2cm}

The SAM segmentation requires prompt inputs to disambiguate the granularity of segmentation. The prompt engineering can be implemented in both fully automated way and by human interaction.

\noindent\textbf{Fully Automated Prompt for Self-Training}: We first densely sample grid points with a spatial distance of 16 pixels as point prompt input. Initial stage segmentation masks $\{\hat{m}_j\}_{j=1\cdots N_{init}}$ are generated by the anchor network with the grid point prompts. We follow the fully automated segmentation introduced by SAM~\cite{kirillov2023segment} to prune out masks with low IoU score and low stability score, followed by NMS suppression. To enable self-training and regularization between different branches, we further generate a fixed set of prompt $\{e_j\}_{j=1\cdots N_{p}}$ from $\{\hat{m}_j\}_{j=1\cdots N_{init}}$ as input to all three branches. As such, the segmentation outputs $\set{Y}^a$, $\set{Y}^s$ and $\set{Y}^t$ are of the same length with exact one-to-one correspondence.

\noindent\textbf{Weak Supervision as Prompt}:
Despite automatic segmentation can be enabled by sampling a grid of point prompts over the image and filtering out low quality and duplicated masks, the quality of segmentation is relatively poor and may contain many false positive predictions, making self-training less effective. Therefore, following prior weakly supervised domain adaptation works~\cite{das2023urban}, we propose to use three types of weak supervisions, including bounding box, sparse point annotation and coarse segmentation mask. Fortunately, in the context of SAM, these weak supervisions perfect match with the prompt inputs which allow seamless integration of weak supervision for adapting SAM.

\subsection{Low-Rank Weights Update}
\vspace{-0.2cm}

The enormous size of backbone network prohibits updating all model weights with large batchsize. However, many existing studies suggest the update of encoder network weights is an effective way of adapting pre-trained models~\cite{liang2021source,su2022revisiting}. To enable updating the backbone network with larger batchsize, we opt for a computation friendly low-rank updating approach~\cite{hu2022lora}. For each weight in the encoder network $\theta\in\mathbbm{R}^{d_i\times d_o}$, we use a low-rank approximation $\omega = AB$ where $A\in\mathbbm{R}^{d_i\times r}$ and $A\in\mathbbm{R}^{r\times d_o}$ with $r$ indicating the rank. We can achieve a compression rate of $r(d_i+d_o)/d_i\cdot d_o$. Only $A$ and $B$ are updated via backpropagation during adaptation to reduce memory footprint. At inference stage, the weight is reconstructed by combining the low rank reconstruction and original weight, $\theta=\theta + AB$.

\section{Experiment}

In this section, we first introduce the experiment settings and downstream tasks for adapting SAM model. We provide detailed comparisons to state-of-the-art methods and qualitative results. Finally, we analyze the effectiveness of individual components and specific designs for the network.

\begin{table*}[!ht]
  \centering
  \caption{Dataset and datasplit used in this work.}
  \vspace{-0.3cm}
    \setlength\tabcolsep{2pt} 
          \resizebox{0.85\linewidth}{!}{
        \begin{tabular}{lr|r|r|r|r|r|r|r|r|r}
    \toprule
          & \multicolumn{2}{c}{Natural Images} & \multicolumn{1}{c}{Corrupted Images} & \multicolumn{2}{c}{Medical Images} & \multicolumn{3}{c}{Camouflaged Objects} & \multicolumn{2}{c}{Robotic Images} \\
    \cmidrule(lr){2-3}\cmidrule(lr){4-4}\cmidrule(lr){5-6}\cmidrule(lr){7-9}\cmidrule(lr){10-11}
          & \multicolumn{1}{c}{COCO~\cite{lin2014microsoft}} & \multicolumn{1}{|c}{Pascal VOC~\cite{everingham2015pascal}} & \multicolumn{1}{|c}{COCO-C} & \multicolumn{1}{|c}{kvasir-SEG~\cite{jha2020kvasir}} & \multicolumn{1}{|c}{ISIC~\cite{gutman2016skin}} & \multicolumn{1}{|c}{CHAMELEON~\cite{skurowski2018animal}} & \multicolumn{1}{|c}{CAMO~\cite{le2019anabranch}} & \multicolumn{1}{|c}{COD10K~\cite{fan2020camouflaged}} & \multicolumn{1}{|c}{OCID~\cite{DBLP:conf/icra/SuchiPFV19}} & \multicolumn{1}{|c}{OSD~\cite{richtsfeld2012segmentation}} \\
\cmidrule{2-11}    \# Training & \multicolumn{1}{|c}{4246} & \multicolumn{1}{|c}{2497} &  \multicolumn{1}{|c}{4246}  &  \multicolumn{1}{|c}{834}  &  \multicolumn{1}{|c}{900}  &    \multicolumn{3}{|c}{4040}  &  \multicolumn{1}{|c}{1972} & \multicolumn{1}{|c}{56} \\
    \# Testing & \multicolumn{1}{|c}{706} &  \multicolumn{1}{|c}{416}  &  \multicolumn{1}{|c}{706}  &  \multicolumn{1}{|c}{166}  &  \multicolumn{1}{|c}{379}  &  \multicolumn{1}{|c}{74}  &  \multicolumn{1}{|c}{250}  &  \multicolumn{1}{|c}{2026}  &  \multicolumn{1}{|c}{328}  & \multicolumn{1}{|c}{55} \\
    \bottomrule
    \end{tabular}%
    }
    \vspace{-0.3cm}
  \label{tab:datasets}%
\end{table*}%

\begin{table*}[t]
\centering
\centering
\caption{\footnotesize{Adaptation results on COCO-C dataset using bounding box prompt.}}
\vspace{-0.3cm}
\resizebox{0.85\linewidth}{!}{
\begin{tabular}{c|c|c|c|c|c|c|c|c|c|c|c|c|c|c|c|c}
\toprule
Method & Brit & Contr & Defoc & Elast & Fog & Frost & Gauss & Glass & Impul & Jpeg & Motn & Pixel & Shot & Snow & Zoom & Avg\\
\midrule
Direct & 72.83 & 57.34 & 64.47 & 69.36 & 72.39 & 70.50 & 67.20 & 64.43 & 67.65 & 68.23 & 62.72 & 68.60 & 67.44 & 69.02 & 58.80 & 66.73 \\
\midrule
TENT & 76.02 & 61.51 & \textbf{67.48} & 70.88 & 74.89 & 73.88 & 69.01 & 67.10 & 69.28 & 70.25 & 65.45 & 70.81 & 69.96 & 72.37 & 62.59 & 69.43\\
SHOT & 73.84 & 59.09 & 65.91 & 69.57 & 73.98 & 72.51 & 68.30 & 66.09 & 68.61 & 69.45 & 64.56 & 70.48 & 68.77 & 71.03 & 60.17 & 68.16\\
Soft Teacher & 73.90 & \textbf{62.12} & 65.41 & 71.32 & 72.16 & 73.27 & 68.84 & 67.49 & 68.73 & 70.18 & \textbf{66.88} & 69.79 & 70.08 & 73.33 & 64.88 & 69.23\\
TRIBE & 76.40 & 60.86 & 66.19 & 72.72 & 75.08 & 75.14 & 70.34 & 66.66 & 70.83 & 72.42 & 65.94 & 70.24 & \textbf{70.66} & 74.22 & 64.56 & 70.15\\
DePT & 69.15 & 57.26 & 59.08 & 66.80 & 58.73 & 66.75 & 66.78 & 62.74 & 65.65 & 66.39 & 61.66 & 66.65 & 67.57 & 66.62 & 58.21 & 64.42\\
WDASS & 76.21 & 60.57 & 67.07 & 72.34 & 75.97 & 74.63 & 69.84 & 67.88 & 69.92 & 71.36 & 66.25 & \textbf{71.99} & 70.32 & 72.25 & 63.61 & 70.01\\
OURS & \textbf{78.50} & 61.05 & 66.99 & \textbf{73.93} & \textbf{77.09} & \textbf{76.10} & \textbf{72.02} & \textbf{68.21} & \textbf{71.29} & \textbf{72.77} & 66.33 & 70.90 & 70.28 & \textbf{75.07} & \textbf{65.33} & \textbf{71.05}\\
\midrule
Supervised & 78.86 & 74.81 & 72.04 & 74.32 & 78.01 & 77.14 & 73.43 & 72.12 & 74.08 & 75.30 & 71.39 & 75.15 & 74.25 & 76.34 & 68.04 & 74.35\\
\bottomrule
\end{tabular}
}
\vspace{-0.5cm}
\label{tab:COCO-C}
\end{table*}

\vspace{-0.2cm}

\subsection{Datasets}
\vspace{-0.2cm}

The source domain training set, SA-1B, was mainly constructed by collecting from natural environments. In this work, we identified five types of downstream segmentation tasks, some of which feature a drastic distribution shift from SA-1B, for evaluation, covering natural clean images, natural corrupted images, medical images, camouflaged images and robotic images. We present the datasets evaluated for each type of downstream task in Tab.~\ref{tab:datasets}.

\noindent\textbf{Datasplit}:
We divide each downstream dataset into non-overlapping training and testing sets following the ratio established by \cite{das2023weakly}. The adaptation step is implemented on the training set. After the model is adapted, we evaluate the model on the held-out testing set. For adaptation to the Camouflaged Objects, following the training protocol in \cite{chen2023sam}, we use the dataset COD10K (the training set of camouflaged images) for training, and use the test set of CAMO, COD10K and the entire CHAMELEON dataset for performance evaluation.

\vspace{-0.2cm}

\subsection{Experiment Details}
\vspace{-0.2cm}
\noindent\textbf{Segment-Anything Model}: We adopt ViT-B~\cite{dosovitskiy2020image} as the encoder network due to memory constraint. The standard prompt encoder and mask decoder are adopted.

\noindent\textbf{Prompt Generation}: We first provide details for generating prompt at both training stages. All training prompts are calculated from ground-truth segmentation mask, simulating the human interactions as weak supervision. Specifically, we extract the minimal bounding box that covers the whole instance segmentation mask as box prompt. The point prompt is created as randomly selecting 5 positive points within ground-truth instance segmentation mask and 5 negative points outside ground-truth mask. The coarse segmentation mask is simulated by fitting a polygon to the ground-truth mask. We choose the number of vertices as $P/20$ with $P$ indicating the perimeter of mask. The minimal number of vertices is 3. We use the same way to generate testing prompt on the testing data. This practice guarantees fair evaluation of SAM model which requires prompt input for segmentation.

\noindent\textbf{Evaluation Metrics}:
We report the mIoU as evaluation metrics. For each input prompt, the IoU is calculated between the ground-truth segmentation mask and predicted mask. The mIoU averages over the IoU of all instances.

\noindent\textbf{Competing Methods}: We evaluate multiple source-free domain adaptation approaches and one state-of-the-art weakly supervised domain adaptive segmentation approach. In specific, direct testing the pre-trained model (\textbf{Direct}) with fixed prompt inputs is prone to the distribution shift and may not perform well on target datasets with significant shift. \textbf{TENT}~\cite{wang2020tent} is a vanilla test-time adaptation method which optimizes an entropy loss for adapting to target domain. \textbf{SHOT}~\cite{liang2021source} employs pseudo label and applies uniform distribution assumption for source-free domain adaptation. \textbf{Soft~Teacher}~\cite{xu2021end} was originally developed for semi-supervised image segmentation. We adapt Soft~Teacher for domain adaptation by keeping the self-training component. \textbf{TRIBE}~\cite{su2023towards} proposed a strong baseline for generic test-time adaptation on continual and class imbalanced domain shifts. We adapt TRIBE for domain adaptive segmentation by replacing the training losses. 
\textbf{DePT}~\cite{gao2022visual} inserts visual prompts into a visual Transformer and adjusts these source-initialized prompts solely during the adaptation process without accessing the source data.
\textbf{WDASS}~\cite{das2023weakly} developed a weakly supervised domain adaptive segmentation method. We also evaluate the upperbound of adapting SAM by fine-tuning with ground-truth segmentation masks, which is dubbed as \textbf{Supervised}. Finally, we evaluate our weakly supervised domain adaptation method as \textbf{Ours}. All the competing methods use the same backbone, i.e. the ViT-B from SAM.

\noindent\textbf{Hyperparameters}: 
We finetune LoRA module of the ViT-B image encoder by Adam optimizer for all experiments. We set the batchsize to 4, distributed over four RTX3090 GPUs, and the learning rate to 0.0001 with a weight decay of 0.0001. For Self-Training loss, we set the hyper-parameters $\gamma$ and $\epsilon$ of $\mathcal{L}^{focal}_{st}$ and $\mathcal{L}^{dice}_{st}$ to 2 and 1 respectively. $\lambda^{focal}_{st}$ is set to 20, which is the coefficients of $\mathcal{L}^{focal}_{st}$. For Anchor loss, the coefficients of two dice losses are denoted as $\lambda^{dice}_{stu}$ and $\lambda^{dice}_{tea}$ , both of which equal 0.5. For Contrast loss, we set the temperature $\tau$ to 0.3. We have set the low rank of the LoRA module for the image encoder to 4.
We apply strong and weak data augmentations for self-training and choices for augmentation follows \cite{xu2021end, sohn2020simple}. 

\subsection{Quantitative Evaluations}
\vspace{-0.2cm}

In this section, we report the quantitative evaluations on the 5 types of target datasets. For each dataset, we also report the adaptation results with weak supervision in the form of bounding box (box), sparse points (point) and coarse segmentation mask (poly).

\noindent\textbf{Adaptation to Corrupted Images}:
Visual corruptions often occur due to sensor fault, bad weather conditions, etc. As seen from Tab.~\ref{tab:COCO-C}, without any adaptation, \textbf{Direct} testing is significantly worse than than the upperbound. With weakly supervisions, all methods could improve over direct testing, in particular, our proposed method outperforms all competing adaptation methods on 10 out of 15 types of corruptions.

\noindent\textbf{Adaptation to Natural Images}:
We then present the results of adapting SAM to natural images in Tab.~\ref{tab:natural_imgs}. For each type of weak supervision, we use the same type of prompt on the testing set. Despite the distribution gap between SA-1B and the target natural images, we observe significant performance gap between \textbf{Supervised} upperbound and \textbf{Direct} baseline, e.g. with box level supervision, there is $5-10\%$ gap in IoU. When weak supervision is provided, both state-of-the-art generic source-free domain adaptation methods and weakly supervised domain adaptive segmentation method improve the generalization on all three types of weak supervisions. Finally, our proposed weakly supervised method achieves a remarkable improvement over all competing methods. The results with box weak supervision is even approaching the fully supervised upperbound.

\begin{table}[h]
\centering
\caption{\footnotesize{Adaptation results on natural clean image datasets.}}
\vspace{-0.3cm}
\resizebox{0.9\linewidth}{!}{
    \begin{tabular}{c|c|c|c|c|c|c}
        \toprule
        \multirow{2}{*}{Method} & 
        \multicolumn{3}{|c}{COCO 2017} &
        \multicolumn{3}{|c}{Pascal VOC} \\
        & box & point & poly & box & point & poly \\
        \midrule
        Direct & 74.29 & 55.06 & 65.64 & 69.21 & 69.21 & 60.79 \\
        \midrule
        TENT & 78.21 & 52.99 & 71.51 & 80.24 & 74.97 & 65.03  \\
        SHOT & 75.18 & 58.46 & 69.26 & 79.80 & 74.26 & 63.38  \\
        Soft Teacher & 75.94 & 43.36 & 68.27 & 72.93 & 56.09 & 62.20 \\
        TRIBE & 77.56 & 49.56 & 70.99 & 78.87 & 69.21 & 65.39 \\
        DePT & 71.00 & 37.35 & 63.27 & 74.09 & 42.99 & 59.94 \\
        WDASS & 77.29 & 60.55 & 70.19 & 80.12 & \textbf{76.15} & \textbf{66.98}  \\
        OURS & \textbf{80.12} & \textbf{62.09} & \textbf{72.33} & \textbf{80.27} & 74.15 & 66.72 \\
        \midrule
        Supervised & 81.50 & 69.77 & 73.39 & 81.23 & 76.98 & 71.32 \\
        \bottomrule
    \end{tabular}
}
\vspace{-0.3cm}
\label{tab:natural_imgs}
\end{table}

\noindent\textbf{Adaptation to Medical Images}: Segmentation for medical images is a major application of foundation models. Our empirical observations in Tab.~\ref{tab:medical} on two medical segmentation datasets suggest that direct applying pre-trained SAM is suboptimal. With weakly supervised adaptation, the segmentation accuracy is greatly improved. We observe the improvement to be particularly significant with point and box which are relatively easy to obtain due to the low labeling cost.

\begin{table}[h]
\centering
\caption{\footnotesize{Adaptation results on medical image segmentation datasets.}}
\vspace{-0.3cm}
\resizebox{0.9\linewidth}{!}{
\begin{tabular}{c|c|c|c|c|c|c}
\toprule
\multirow{2}{*}{Method} & 
\multicolumn{3}{|c}{kvasir-SEG} &
\multicolumn{3}{|c}{ISIC} \\
& box & point & poly & box & point & poly \\
\midrule
Direct & 81.59 & 62.30 & 54.03 & 66.74 & 53.42 & 62.82 \\
\midrule
TENT & 82.47 & 61.84 & 62.97 & 71.76 & 53.46 & 67.12 \\
SHOT & 82.30 & 63.76 & 61.34 & 71.99 & 55.99 & 66.86 \\
Soft Teacher & 84.12 & 73.53 & 58.15 & 75.74 & 54.95 & 72.29 \\
TRIBE & 85.05 & 73.03 & 64.61 & 72.61 & 50.36 & 67.99 \\
DePT & 81.91 & 52.06 & 61.55 & 78.43 & 46.79 & 72.75 \\
WDASS & 84.01 & 63.78 & 64.78 & 74.23 & 55.63 & 67.84 \\
OURS & \textbf{85.47} & \textbf{75.23} & \textbf{67.40} & \textbf{80.01} & \textbf{62.12} & \textbf{75.36} \\
\midrule
Supervised & 85.89 & 77.54 & 81.64 & 81.62 & 79.81 & 80.26 \\
\bottomrule
\end{tabular}
}
\vspace{-0.5cm}
\label{tab:medical}
\end{table}

\noindent\textbf{Adaptation to Camouflaged Objects}: We further evaluate on adapting SAM to camouflaged object detection. We evaluate on three camouflaged object detection datasets with results reported in Tab.~\ref{tab:camouflage}. Preliminary studies revealed that SAM is particularly vulnerable to camouflaged objects~\cite{tang2023can} due to the low contrast between background and foreground. We make similar observations here that even with the help of prompt inputs at testing stage, \textbf{Direct} inference suffers a lot, achieving less than 66\% mIoU. When weakly labeled data are used for adaptation, we observe very significant improvements, with point prompt only we improved by a staggering 35\% in mIoU on the CHAMELEON dataset. Our proposed method is also consistently better than all competing methods on all types of weak supervisions.

\begin{table}[h]
\centering
\caption{\footnotesize{Adaptation results on camouflaged object datasets.}}
\vspace{-0.3cm}
\resizebox{0.9\linewidth}{!}{
\begin{tabular}{c|c|c|c|c|c|c|c|c|c}
\toprule
\multirow{2}{*}{Method} & 
\multicolumn{3}{|c}{CHAMELEON} &
\multicolumn{3}{|c}{CAMO} &
\multicolumn{3}{|c}{COD10K} \\
& box & point & poly & box & point & poly & box & point & poly \\
\midrule
Direct & 51.32 & 39.37 & 45.78 & 62.72 & 57.43 & 50.85 & 66.32 & 63.61 & 40.04 \\
\midrule
TENT & 65.48 & 54.53 & 53.06 & 71.24 & 59.59 & 60.29 & 69.36 & 61.94 & 43.36 \\
SHOT & 68.60 & 62.47 & 54.36 & 71.61 & 62.78 & 58.72 & 69.09 & 65.25 & 42.38\\
Soft Teacher & 65.92 & 44.17 & 46.72 & 62.30 & 48.64 & 51.26 & 66.32 & 50.04 & 32.27\\
TRIBE & 71.00 & 52.80 & 54.99 & 66.00 & 61.97 & 60.54 & 67.84 & 63.62 & 42.75 \\
DePT & 54.48 & 33.46 & 42.47 & 55.44 & 33.07 & 48.63 & 59.32 & 34.06 & 35.51\\
WDASS & 71.91 & 62.40 & 56.80 & 71.25 & 63.39 & 62.29 & 71.42 & 65.61 & 43.93 \\
OURS & \textbf{75.94} & \textbf{74.00} & \textbf{66.83} & \textbf{73.42} & \textbf{65.55} & \textbf{62.90} & \textbf{71.93} & \textbf{70.55} & \textbf{45.87} \\
\midrule
Supervised & 78.05 & 85.86 & 68.38 & 79.17 & 77.01 & 67.12 & 78.06 & 78.44 & 64.90 \\
\bottomrule
\end{tabular}
}
\vspace{-0.5cm}
\label{tab:camouflage}
\end{table}

\noindent\textbf{Adaptation to Robotic Images}: Finally, we evaluate adapting SAM to image collected from robotic environments where segmenting out the spatial extent of individual objects is essential for interactions. As seen from the results in Tab.~\ref{tab:robotic}, adapting SAM to robotic images with weak supervision further increase the segmentation accuracy. The improvement is more noticeable with point and polygon weak supervisions.

\begin{table}[h]
\centering
\vspace{-0.3cm}
\caption{\footnotesize{Adaptation results on robotic image datasets.}}
\vspace{-0.3cm}
\resizebox{0.9\linewidth}{!}{
\begin{tabular}{c|c|c|c|c|c|c}
\toprule
\multirow{2}{*}{Method} & 
\multicolumn{3}{|c}{OCID} &
\multicolumn{3}{|c}{OSD} \\
& box & point & poly & box & point & poly \\
\midrule
Direct & 86.35 & 71.41 & 72.81 & 87.62 & 78.86 & 80.77 \\
\midrule
TENT & 87.77 & 66.61 & \textbf{77.53} & 88.10 & 80.53 & 87.85 \\
SHOT & 88.06 & 74.39 & 76.25 & 88.09 & 80.52 & 87.86 \\
Soft Teacher & 84.98 & 68.46 & 73.75 & 90.41 & 80.49 & 87.00 \\
TRIBE & 86.77 & 67.86 & 76.50 & 90.42 & \textbf{80.54} & 87.84 \\
DePT & 82.00 & 56.52 & 70.92 & 81.84 & 69.06 & 82.50 \\
WDASS & 87.68 & 77.13 & 76.70 & 88.07 & 80.52 & 88.19 \\
OURS & \textbf{88.09} & \textbf{80.14} & 77.41 & \textbf{92.11} & 80.51 & \textbf{89.72} \\
\midrule
Supervised & 91.24 & 89.22 & 79.23 & 92.14 & 82.41 & 90.83 \\
\bottomrule
\end{tabular}
}
\vspace{-0.3cm}
\label{tab:robotic}
\end{table}

\subsection{Qualitative Evaluations}
\vspace{-0.2cm}

\begin{figure*}[!htb]
    \centering
\includegraphics[width=0.9\linewidth]{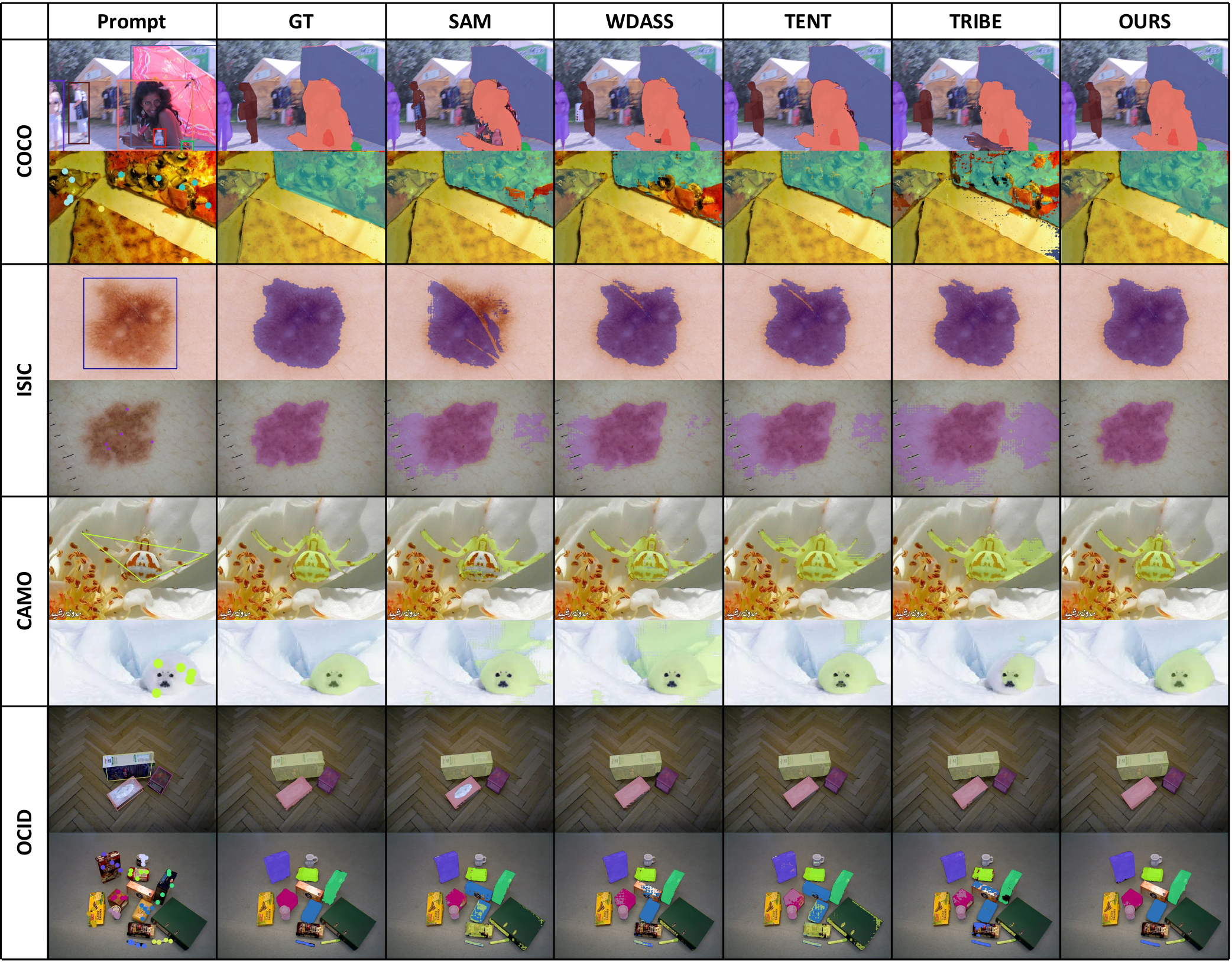}
\vspace{-0.2cm}
    \caption{qualitative results on some selected examples. Three types of prompts at testing stage are visualized for reference.}
    \vspace{-0.4cm}
    \label{fig:VISUAL}
\end{figure*}

We present the qualitative comparisons of segmentation results for selected examples in Fig.~\ref{fig:VISUAL}. We make the following observations from the results. Evaluating pre-trained SAM (SAM) on COCO with both box and point prompt exhibits very fine-grained segmentation results. However, the model is overly sensitive to the high contrast regions, e.g. the hair for the girl and the toppings of pizza are missing from the segmentation mask. With the proposed weakly supervised adaptation approach, our method is able to generate more smoothed segmentation mask and the mask better reflects the semantic boundary. Similar observations are made from the medical image segmentation dataset. Without adapting the SAM model, segmentation results are either too conservative with low recall or expanding beyond the semantic boundary. With our weakly supervised adaptation, SAM is able to produce high fidelity results. We further present results on camouflaged segmentation task (CAMO). Without adaptation, SAM fails to segment out the spider's belly and confuses the snow background with the harp seal in the foreground. Our weakly adapted SAM successfully segments out the spider's belly and produces high quality mask for the harp seal. Finally, two examples from OCID dataset suggest that the pre-trained SAM tends to over-segment, e.g. missing the open whole for the tissue box, or confusing adjacent objects, e.g. merging the pen with the folder label. With weakly supervised adaptation SAM picks up the semantic meaning and produces segmentation results that better aligns with human intention.

\subsection{Ablation Study}
\vspace{-0.2cm}

In this section, we analyze the effectiveness of individual components on COCO dataset. As presented in Tab.~\ref{tab:ablation}, when self-training is applied alone, we observe a significant performance drop ($74.29\%\rightarrow58.88\%$ with box weak supervision), suggesting the severity of confirmation bias. This issue can be well remedied when the anchor loss regularization is applied. Self-training with anchored regularization already improves the performance of SAM after adaptation. Finally, the contrastive loss directly regularizes the encoder network's output and contributes with additional improvement of segmentation accuracy. We also investigate the final self-training architecture without any weak supervision, i.e. the pseudo label masks are generated with grid point prompt. The results after adaptation is slightly worse than with weak supervision but consistently better than without adaptation on both box and coarse mask testing prompts. Nevertheless, the results with point prompt performs even worse than without adaptation, suggesting the necessity of weakly supervised adaptation.

\begin{table}[t]
\centering
\caption{\footnotesize{Ablation studies of the proposed weakly supervised adaptation method on COCO dataset.}}
\vspace{-0.3cm}
\resizebox{0.9\linewidth}{!}{
\begin{tabular}{c|c|c|c|c|c|c}
\toprule
Weak~Sup. & Self-Train. & Anchor & Cont. Loss & box & point & poly\\
\midrule
\multicolumn{4}{c|}{original SAM} & 74.29 & 56.36 & 65.42 \\
\checkmark & \checkmark &  &  &  58.88 & 32.51 & 55.03 \\
\checkmark & \checkmark & \checkmark &  & 79.65 & 57.25 & 70.49 \\
\checkmark & \checkmark &  & \checkmark & 62.95 & 22.87 & 56.91 \\
\checkmark & \checkmark & \checkmark & \checkmark & 80.12 & 62.09 & 72.33 \\
\midrule
 & \checkmark & \checkmark & \checkmark & 76.18 & 47.63 & 70.44 \\
\bottomrule
\end{tabular}
}
\vspace{-0.3cm}
\label{tab:ablation}
\end{table}

\subsection{Additional Analysis}
\vspace{-0.2cm}

\noindent\textbf{Generalization to Cross-Prompt Testing}: We further investigate the effectiveness of adaptation by testing with all three types of prompts. Specifically, we compare with the best~(WDASS) and second best~(TENT) methods with results presented in Tab.~\ref{tab:CrossPrompt}. Apart from using point as testing prompt, our methods consistently improves the performance even though the testing prompt is different from training weak supervision.

\begin{table}[t]
\centering
\caption{\footnotesize{Adaptation results on cross-prompt testing scenarios.}}
\vspace{-0.3cm}
\resizebox{0.9\linewidth}{!}{
\begin{tabular}{c|c|c|c|c|c|c|c|c|c}
\toprule
\multicolumn{1}{c}{Tr. Weak. Sup.} & 
\multicolumn{3}{|c}{box} &
\multicolumn{3}{|c}{point} &
\multicolumn{3}{|c}{polygon} \\
\cmidrule(lr){2-4}\cmidrule(lr){5-7}\cmidrule(lr){8-10}
Te. Prompt & box & point & poly & box & point & poly & box & point & poly \\
\midrule
Direct & 74.29 & 54.76 & 66.75 & 74.29 & 54.64 & 66.75 & 74.29 & 54.76 & 65.64 \\
TENT & 78.21 & 54.83 & 68.82 & 52.99 & 52.99 & 68.58 & 77.08 & 53.90 & 71.51 \\
WDASS & 77.29 & \textbf{58.78} & 68.94 & 74.69 & 60.55 & 69.15 & 75.27 & \textbf{55.27} & 70.19 \\
\midrule
OURS & \textbf{80.12} & 52.28 & \textbf{72.07} & \textbf{76.49} & \textbf{62.09} & \textbf{71.02} & \textbf{78.57} & 52.88 & \textbf{72.33} \\
\bottomrule
\end{tabular}
}
\vspace{-0.5cm}
\label{tab:CrossPrompt}
\end{table}

\noindent\textbf{Effectiveness of Updating Different Components}: There are many options for updating the pre-trained SAM model. Given limited computing resources, choosing the appropriate components of network to update is crucial to optimize the generalization performance. In this section, we investigate several options of network components to update during adaptation. \textbf{Full finetune} means finetuning the whole encoder. \textbf{MaskDecoder} finetune the whole decoder of SAM without additional redundant learnable parameters. \textbf{LayerNorm} finetune the all layernorms of the SAM image encoder. \textbf{LoRA} finetunes the encoder network in through low-rank decomposition only. \textbf{EVP} use the Embedding Tune and the HFC Tune to tune the extracted features by image encoder.
We present the investigation on the COCO dataset with bounding boxes as weak supervision in Tab.~\ref{tab:DifferComponent}.
In particular, we have also explored the impact of different combinations of components, e.g. MaskDecoder + LoRA, LoRA + EVP, and so on.
The results suggest LoRA finetuning the encoder network alone yields the best performance.

\begin{table}[t]
\centering
\caption{\footnotesize{Adaptation results produced by finetuning different modules.}}
\vspace{-0.3cm}
\resizebox{0.65\linewidth}{!}{
\begin{tabular}{c|c}
\toprule
Finetuning Module & IoU  \\
\midrule
None & 74.30  \\
Full finetune & 78.54  \\
MaskDecoder & 76.81 \\
LayerNorm & 79.30 \\
Low-Rank Adaptation (LoRA) & \textbf{80.12} \\
Explicit Visual Prompt (EVP) & 77.38 \\
MaskDecoder + LoRA & 77.28 \\
MaskDecoder + EVP & 76.83 \\
LoRA + EVP & 79.47 \\
LoRA + LayerNorm & 79.35 \\
MaskDecoder + LoRA + EVP & 77.31 \\
\bottomrule
\end{tabular}
}
\vspace{-0.5cm}
\label{tab:DifferComponent}
\end{table}

\vspace{-0.06cm}

\section{Conclusion}
\vspace{-0.2cm}

We studies the generalization of Segment-Anything model for multiple downstream image segmentation tasks. A tangible solution with no access to source dataset and low memory cost was proposed to adapt SAM to downstream data in a source-free domain adaptation manner. The proposed method is naturally compatible with weak supervisions which could substantially improve the efficacy of adaptation. Extensive evaluations on 10 datasets from 5 types of downstream tasks suggest the proposed adaptation method can significantly improve the generalization of SAM under various degrees of distribution shift.

\noindent\textbf{Acknowledgements.} This work is supported by National Natural Science Foundation of China (NSFC) (Grant Number: 62106078), and Agency for Science, Technology and Research (Grant Number: C210112059). This work was partially done during Yongyi Su's attachment with Institute for Infocomm Research (I2R), funded by China Scholarship Council~(CSC).

{
    \small
    \bibliographystyle{ieeenat_fullname}
    \bibliography{main}
}

\newpage

\appendix

\section*{Appendix}

In this supplementary material, we first investigate the impact of the amount of weak supervision on adaptation performance. We further demonstrate that after weakly supervised adaptation we can further improve the performance of one-shot segmentation. Additional analysis of hyper-parameters and qualitative results are presented as well.

\section*{Table of contents:}

\begin{itemize}[leftmargin=0.8cm]
    \item \S\ref{sec:A}: The Impact of Weak Label Numbers
    \item \S\ref{sec:B}: One-Shot Personalized Segmentation
    \item \S\ref{sec:C}: Hyper-Parameter Sensitivity
    \item \S\ref{sec:D}: Additional Experiments
    \item \S\ref{sec:E}: Visualization Examples
\end{itemize}

\section{The Impact of Weak Label Numbers on Performance}\label{sec:A}

\begin{figure*}[h]
    \centering
\vspace{0.2cm}
\subfloat[][Using box as testing prompt.]{\includegraphics[width=0.33\linewidth]{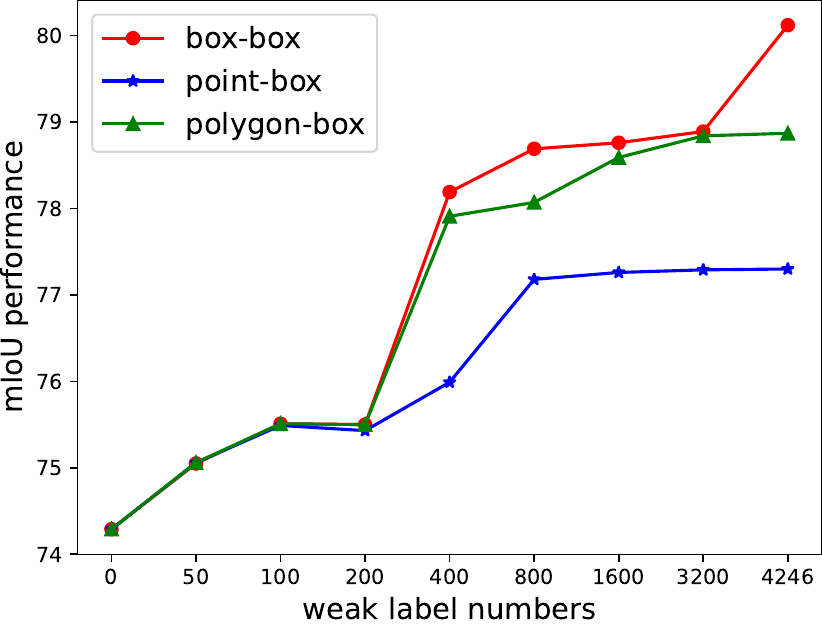}}
\subfloat[][Using point as testing prompt.]{
\includegraphics[width=0.33\linewidth]{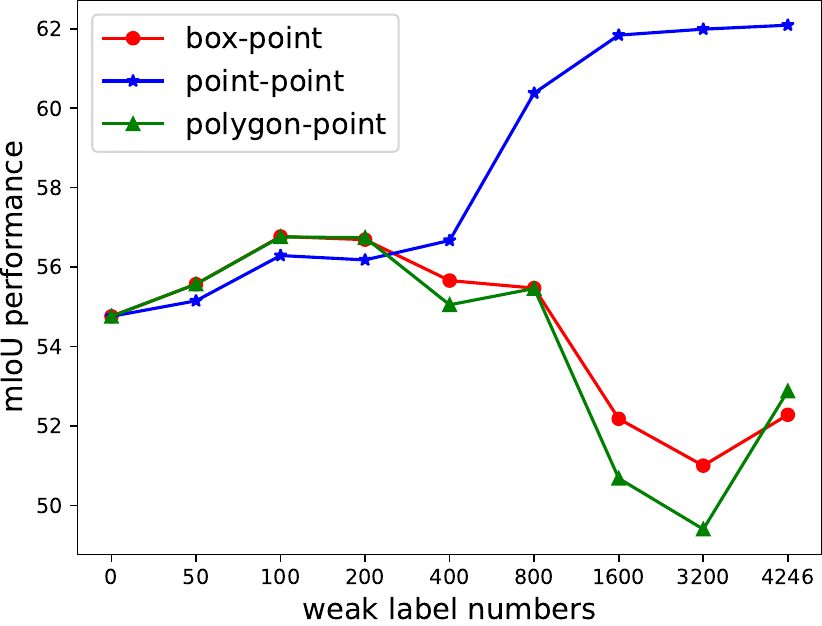}}
\subfloat[][Using polygon as testing prompt.]{
\includegraphics[width=0.33\linewidth]{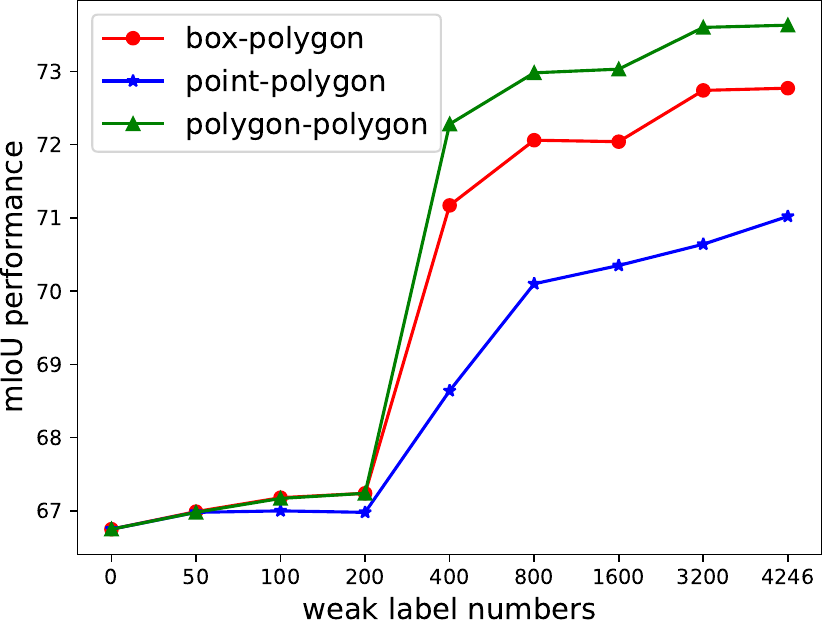}}
    \caption{Annotation cost vs. performance. 1. Performance of different numbers of weak labels on performance. 2. Performance of three weak labels under the same prompt verification.}
    \vspace{-0.4cm}
    \label{fig:box}
\end{figure*}

In this experiment, we aim to demonstrate the cost-effectiveness of utilizing weak labels. To facilitate comparison, we incrementally choose weakly labeled images of 50, 100, 200, 400, 800, 1600, 3200, and 4246 in COCO dataset for adaptation. We adapt the model using three different types of weak labels and evaluate their performance using three different prompts. We make the following observations from Fig.~\ref{fig:box}~(a-c). First, when training/adapation weak supervision is the same with testing prompt, we observe the most effective generalization of SAM. Moreover, the generalization improves upon more weak supervision except for adaptation with point label and testing with box and polygon prompts. This suggest the mask decoder is still sensitive to the shift of prompt used for training and testing.


\section{One-Shot Personalized Segmentation}\label{sec:B}

We further investigate the effectiveness of adapting SAM for one-shot personalized segmentation. Specifically, PerSAM~\cite{zhang2023personalize} is a training-free personalization approach for SAM that enables it to achieve one-shot segmentation based on visual cues. Given only a single image with a reference mask, PerSAM first localizes the target concept by a location prior in the test image, and then segments the target object through target-guided attention, target-semantic prompting, and cascaded post-refinement. 

In the main text, we have demonstrated that SAM performs poorly when facing significant domain shift. Similarly, in one-shot tasks, there may also be downstream tasks with significant domain shift. To demonstrate the effectiveness of weakly supervised adaptation, we use the PerSAM framework and conduct one-shot experiments on the ISIC dataset for medical image segmentation. In particular, we compare three alternative designs for PerSAM. First, we directly use PerSAM with one-shot reference image and test on the testing set of ISIC. We further evaluate PerSAM-F which finetunes the mask weight on one-shot reference image. Finally, our method adapts SAM with weakly labeled data and then use the adapted SAM for one-shot PerSAM. We further improved PerSAM by sampling 30 feature points on the reference image using Kmeans to diversify the reference prototypes. The reference prototypes are used to query the positive point on the testing image as prompt for SAM. We denot the improved version as PerSAM$^*$. 

\begin{figure*}
    \centering
\vspace{0.2cm}
\includegraphics[width=0.9\linewidth]{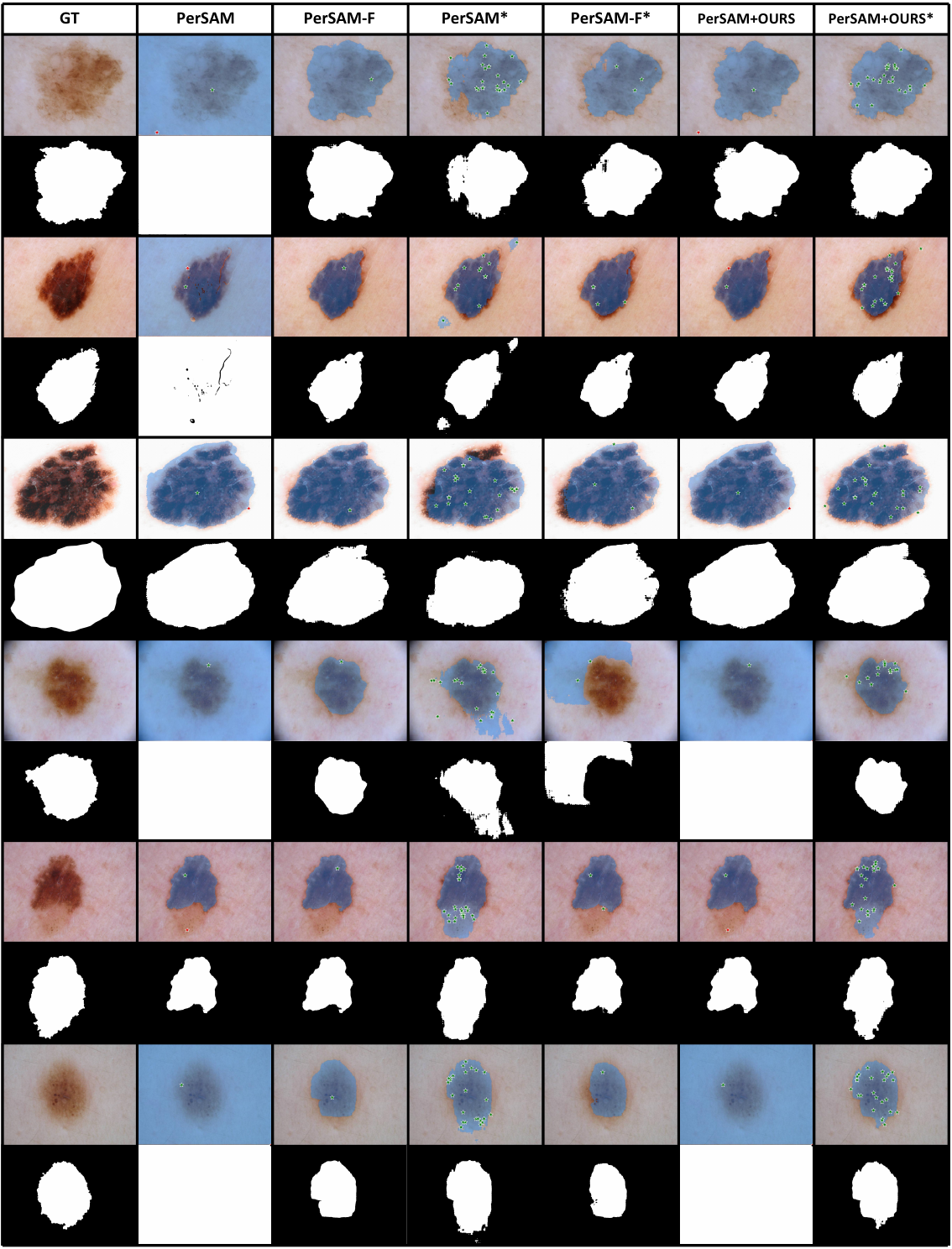}
    \caption{Qualitative results for One-Shot PerSAM segmentation.}
    \label{fig:oneshot}
\end{figure*}

As shown in Tab.~\ref{tab:OneShot}, we observe that without any modifications to PerSAM, our adapted SAM achieves better one-shot segmentation performance than the original PerSAM model in terms of both IoU and Accuracy~(Acc). With the improved PerSAM$^*$, our weakly supervised adaptation is much more effective in localizing the target objects outperforming both the original PerSAM and the improved PerSAM with 6-10$\%$ IoU. We also visualize the one-shot segmentation results in Fig.~\ref{fig:oneshot}. The green and red stars refer to the positive and negative point prompt on the testing image. The original PerSAM tends to either ignore the foreground partially or over estimate the foreground. While our improved PerSAM$^*$, thanks to the multiple point prompt, achieves better localization of the foreground object.

The above experiment was conducted on medical image segmentation with a large domain gap, e.g. ISIC Dataset. To enhance the persuasiveness, especially that our method improves model generalization, not just task-specific adaptation. We supplemented experiments on one-shot segmentation in natural images in Tab.~\ref{tab:OneShot-COCO}. We observe that i) there is a clear improvement with mask mixture finetuning (PerSAM-F); and ii) PerSAM + OURS is still better than PerSAM-F. The improvement of PerSAM-F suggest the segmentation granularity is still a concern for applying SAM for in-distribution downstream task. When specific information on segmentation target is considered (PerSAM + OURS) the performance is substantially better (31.89 v.s. 21.22 in mIoU). This suggests segmentation specific information (e.g. spatial extent) is also crucial for adapting SAM to in-distribution downstream tasks.


\begin{table}[t]
\centering
\caption{\footnotesize{Oneshot experimental results on the ISIC dataset. * indicates that the reference prototypes are collected by Kmeans algorithm.}}
\resizebox{0.5\linewidth}{!}{
\begin{tabular}{c|c|c}
\toprule
Method & IoU & Acc \\
\midrule
PerSAM & 38.16 & 87.48 \\
PerSAM-F & \textbf{41.06} & 82.73  \\
PerSAM~+~OURS & 40.00 & \textbf{87.93} \\
\midrule
PerSAM* & 43.96 & 70.57 \\
PerSAM-F* & 40.97 & 57.98 \\
PerSAM~+~OURS* & \textbf{49.86} & \textbf{72.84} \\
\bottomrule
\end{tabular}
}
\vspace{-0.3cm}
\label{tab:OneShot}
\end{table}

\begin{table}[t]
\centering
\caption{\footnotesize{Oneshot segmentation results on COCO dataset. 
}}
\resizebox{0.8\linewidth}{!}{
    \begin{tabular}{l|c|c}
        \toprule
        Method & mIoU & Acc \\
        \midrule
        PerSAM & 21.22 & 32.33 \\
        PerSAM-F & 23.43 & 47.86  \\
        PerSAM~+~OURS (w/ box weak supervision) & \textbf{31.89} & \textbf{70.55} \\
        \bottomrule
    \end{tabular}
}
\vspace{-0.3cm}
\label{tab:OneShot-COCO}
\end{table}

\section{Hyper-Parameter Sensitivity}\label{sec:C}

In this section, we evaluate the sensitivity to different hyper-parameters.
For Anchor loss, the coefficients of two dice losses are denoted as $\lambda^{dice}_{stu}$ and $\lambda^{dice}_{tea}$ , For the Anchor loss, the coefficients of the two dice losses are denoted as $\lambda^{dice}_{stu}$ and $\lambda^{dice}_{tea}$, respectively, and are set as follows: 1.0:0, 0.7:0.3, 0.5:0.5, 0.3:0.7, 0:1.0. For Contrast loss, we set the temperature $\tau$ to 0.1, 0.3, 0.5. For model finetuning, we use the Adam optimizer with learning rates set to 0.001, 0.0001, and 0.00001, respectively. As shown in Tab~\ref{tab:H-Param}, out proposed weakly supervised adaptation method is relatively stable to the choice of hyper-parameters.

\begin{table}[t]
\centering
\caption{\footnotesize{Experimental results of Hyper-Parameter sensitivity analysis on COCO dataset. The bold text indicates which values are used in our method.}}
\vspace{-0.3cm}
\resizebox{0.65\linewidth}{!}{
\begin{tabular}{c|c|c|c|c}
\toprule
 \multicolumn{2}{c|}{Hyper-Param.} & box & point & poly\\
 \midrule
 \multirow{3}{*}{\rotatebox{90}{Temp. $\tau$}}
 & $0.1$ & 79.05 & 63.65 & 72.98 \\
 & $0.3$ & \textbf{79.89} & \textbf{63.86} & \textbf{73.01} \\
 & $0.5$ & 79.19 & 63.46 & 72.96 \\
 \midrule
 \multirow{3}{*}{\rotatebox{90}{LR}}
 & $1e-3$ & 78.42 & \textbf{63.90} & 72.91 \\
 & $1e-4$ & \textbf{79.89} & 63.87 & \textbf{73.01} \\
 & $1e-5$ & 79.01 & 63.81 & 72.99 \\
 \midrule
 \multirow{5}{*}{\rotatebox{90}{$\lambda^{dice}_{stu} : \lambda^{dice}_{tea}$}}
 & $1.0 : 0.0$ & 67.19 & 41.64 & 64.4 \\
 & $0.7 : 0.3$ & 72.98 & 42.11 & 68.62 \\
 & $0.5 : 0.5$ & \textbf{79.89} & \textbf{63.82} & \textbf{73.01} \\
 & $0.3 : 0.7$ & 79.76 & 63.60 & 72.68 \\
 & $0.0 : 1.0$ & 79.26 & 63.16 & 72.62 \\
\bottomrule
\end{tabular}
}
\vspace{-0.3cm}
\label{tab:H-Param}
\end{table}

\section{Additional Experiments}\label{sec:D}

\textbf{Preferring ``Shared Weights'' over ``EMA''} In our approach, the teacher and student models share the same weights. Another approach is using EMA for teacher model weights. Comparative experiments with "EMA weight" in Tab.~\ref{tab:table-add} show weight sharing's superior performance.

\noindent\textbf{EWC regularization} Regularizing the model weights is subject to the difference in scale and size of model weights. We adopt anchor regularization in self-training. For other regularization methods, such as Elastic Weight Consolidation\cite{kirkpatrick2017overcoming}, we adapt this approach for SAM adaptation, and the results ``EWC reg.'' in Tab.~\ref{tab:table-add} suggest our proposed regularization is still optimal.

\begin{table}[!h]
    \centering
    \vspace{-0.3cm}
    \caption{The additional experiments}
    \vspace{-0.3cm}
    \resizebox{0.9\linewidth}{!}{
        \begin{tabular}{c|c|c|c|c|c|c}
        \toprule[1.2pt]
            \multirow{2}{*}{Method} & \multicolumn{3}{c|}{COCO 2017} & \multicolumn{3}{c}{ISIC} \\
            & box & point & poly & box & point & poly \\
        \midrule
            Direct & 74.29 & 55.06 & 65.64 & 66.74 & 53.42 & 62.82 \\
            EMA weights & 78.14 & 55.03 & 73.22 & 78.12 & 63.41 & 73.74  \\
            EWC reg. & 76.44 & 52.52 & 71.33 & 78.87 & \textbf{66.89} & 75.88 \\
            OURS & \textbf{80.12} & \textbf{64.39} & \textbf{73.72} & \textbf{80.26} & 63.90 & \textbf{76.59} \\
        \bottomrule[1.2pt]
        \end{tabular}
    }
    \vspace{-0.3cm}
    \label{tab:table-add}
\end{table}

\section{Visualization Examples on Various Downstream Domains}\label{sec:E}

\begin{figure*}
    \centering
    \includegraphics[width=0.9\linewidth]{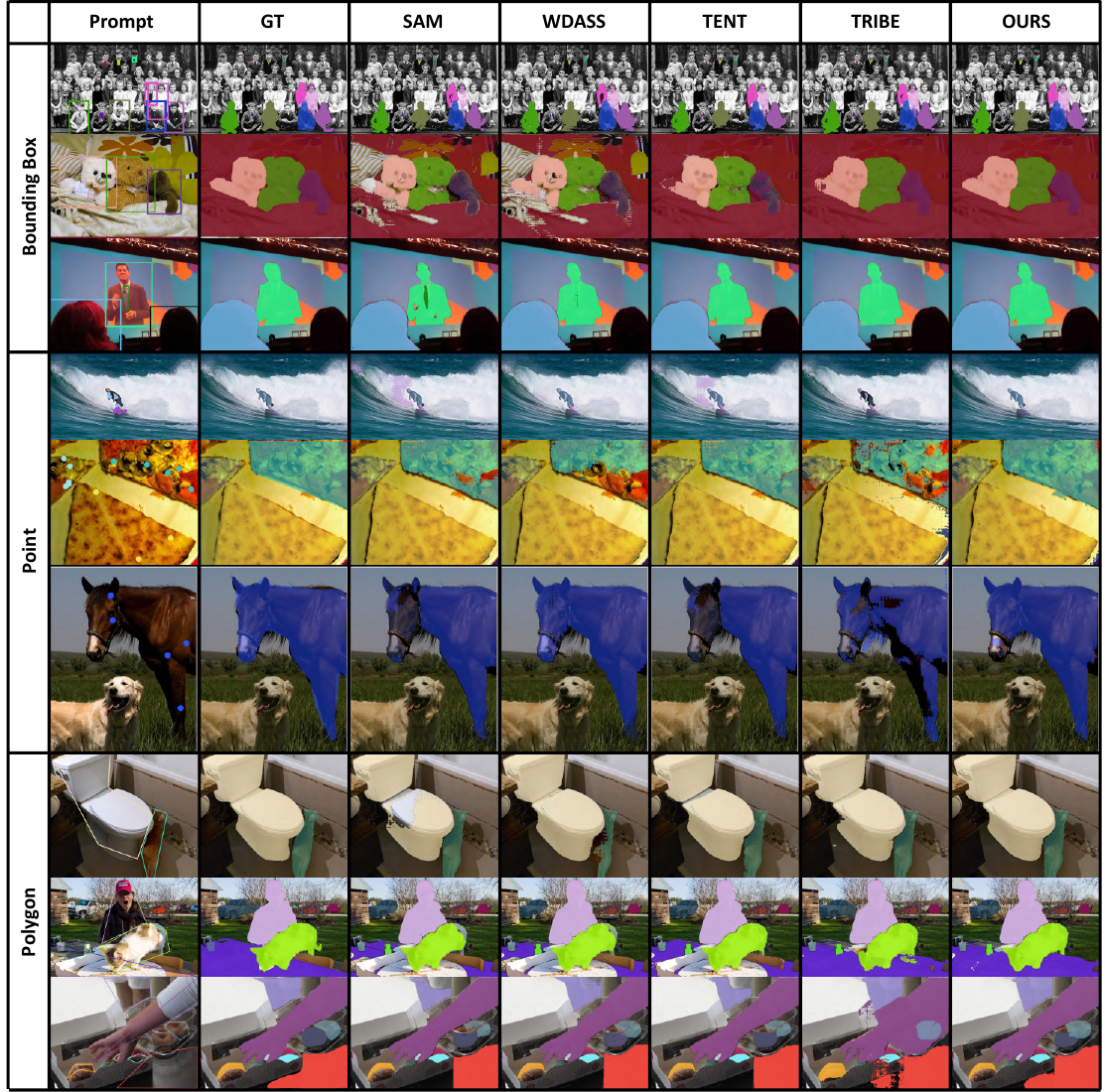}
    \caption{Visualization examples on COCO dataset.}
    \label{fig:vis_coco}
\end{figure*}

\begin{figure*}
    \centering
    \includegraphics[width=0.9\linewidth]{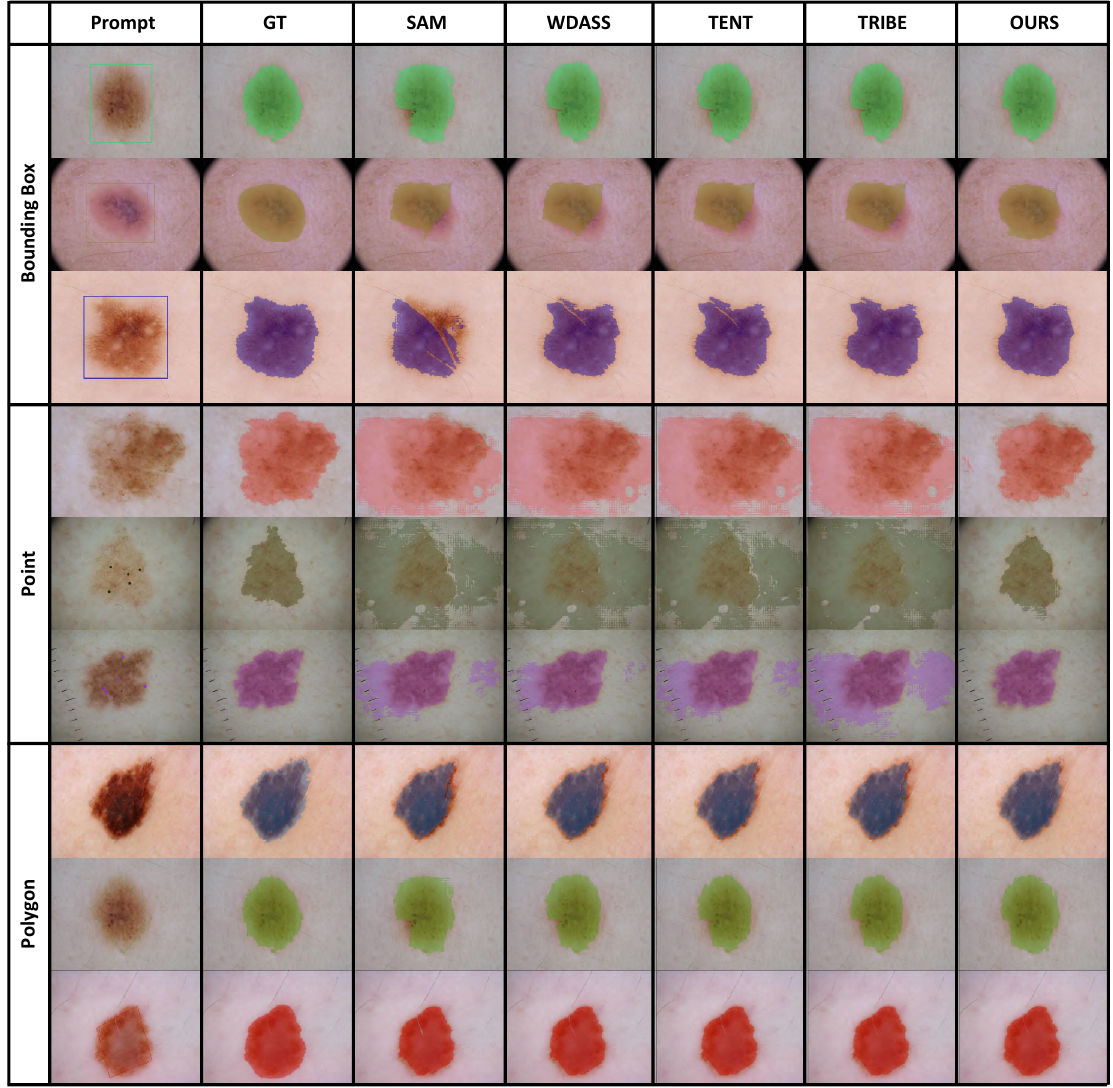}
    \caption{Visualization examples on ISIC dataset.}
    \label{fig:vis_isic}
\end{figure*}

\begin{figure*}
    \centering
    \includegraphics[width=0.9\linewidth]{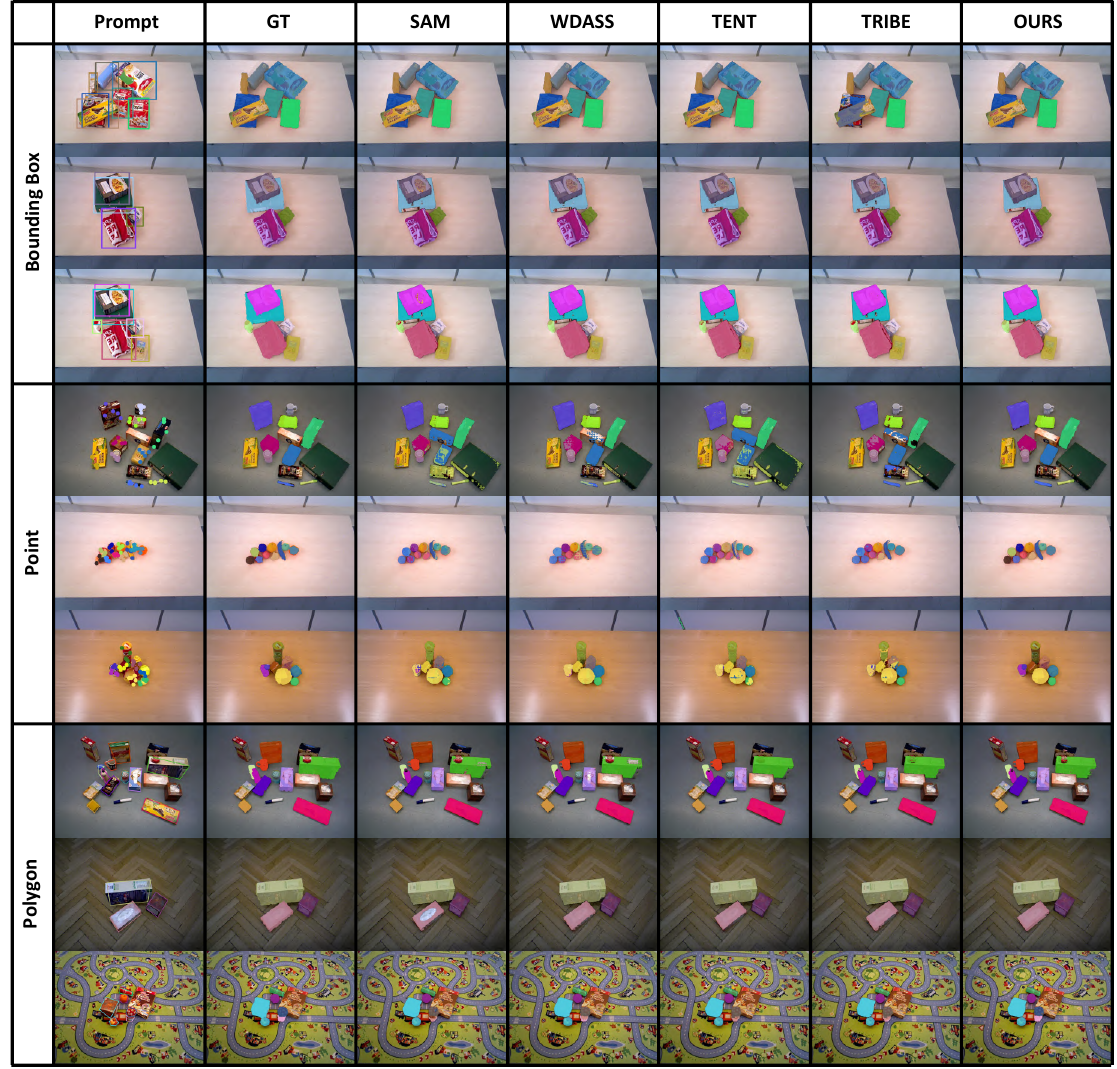}
    \caption{Visualization examples on OCID dataset.}
    \label{fig:vis_ocid}
\end{figure*}

\begin{figure*}
    \centering
    \includegraphics[width=0.9\linewidth]{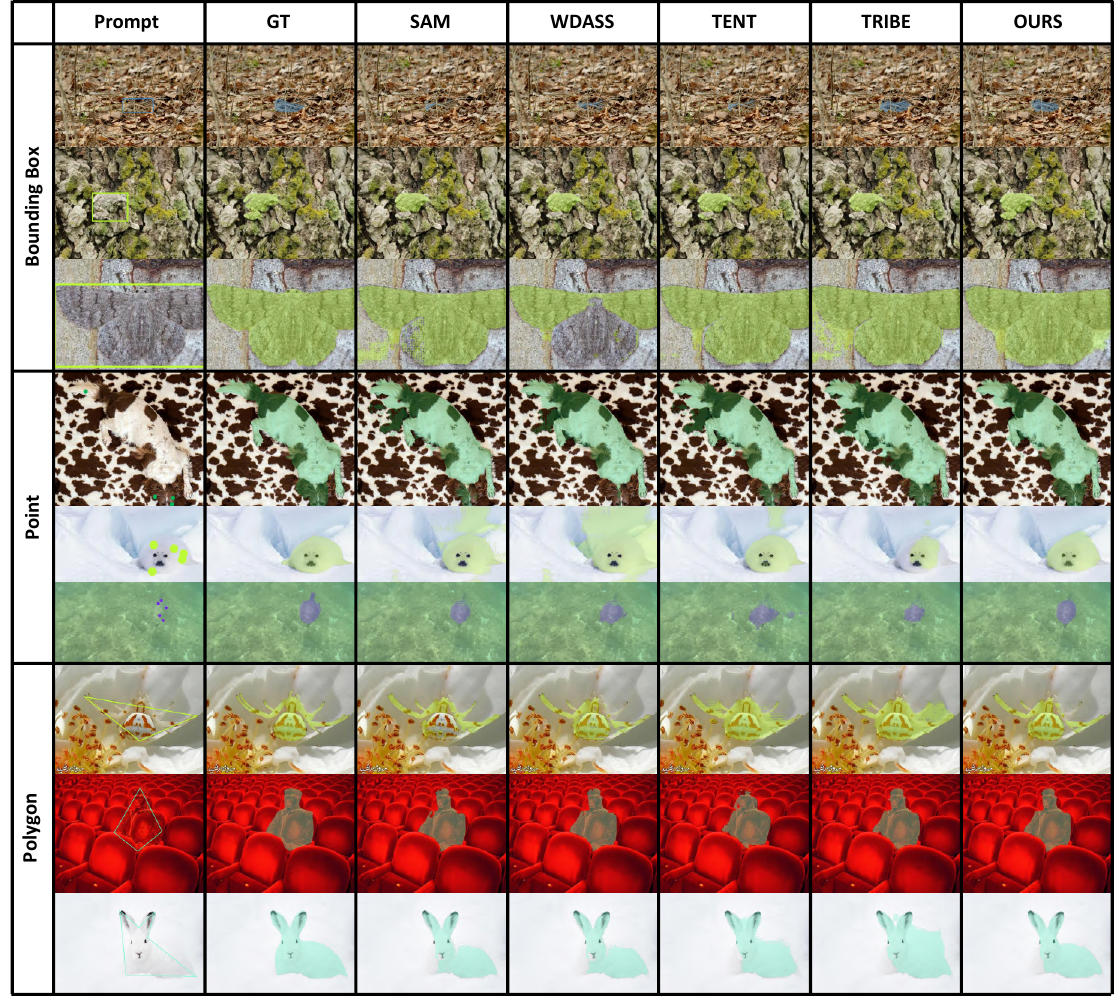}
    \caption{Visualization examples on CAMO dataset.}
    \label{fig:vis_camo}
\end{figure*}

\begin{figure*}
    \centering
    \includegraphics[width=0.9\linewidth]{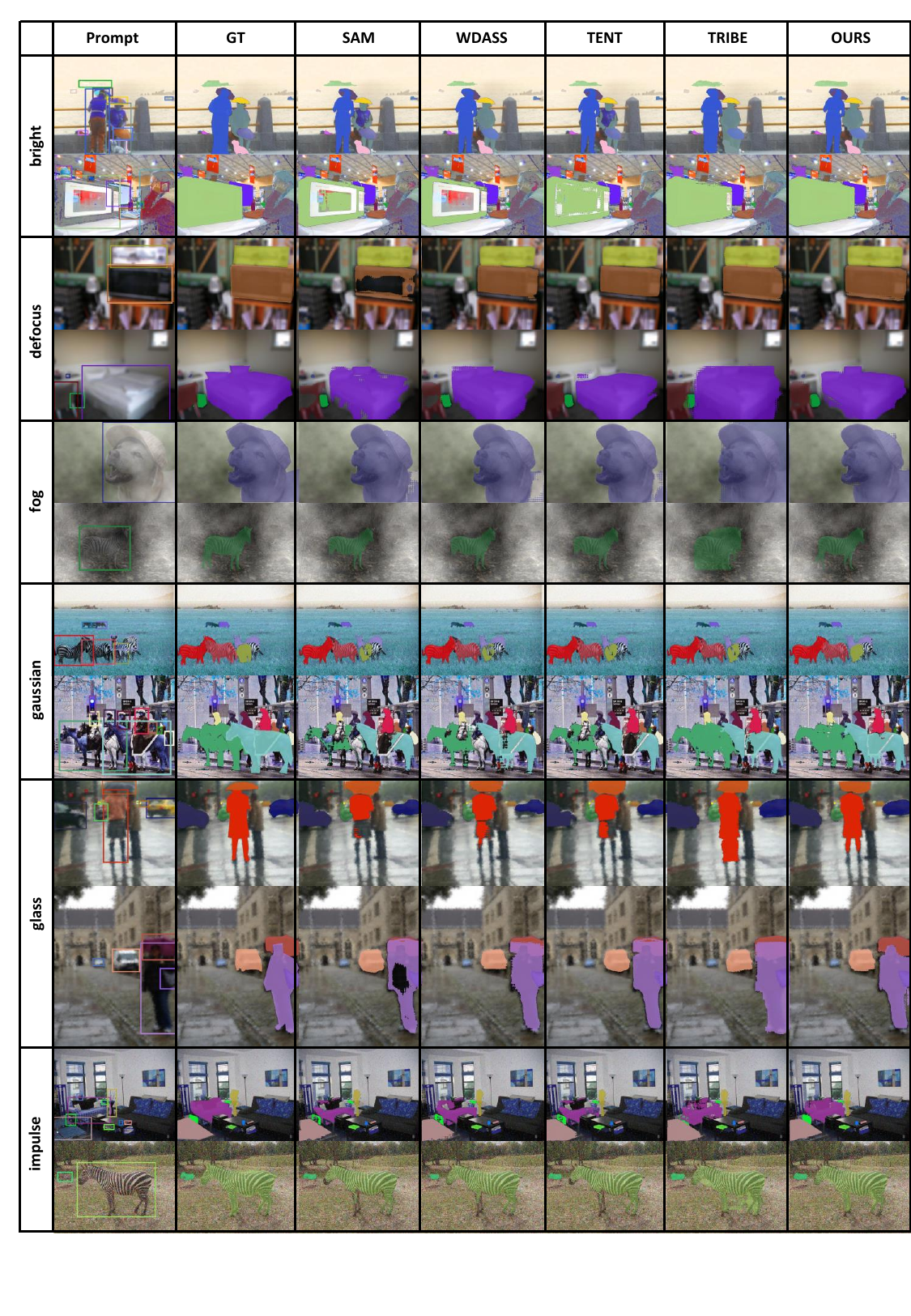}
    \vspace{-1.2cm}
    \caption{Visualization examples on COCO-C dataset.}
    \label{fig:corrupt-1}
\end{figure*}

\begin{figure*}
    \centering
    \includegraphics[width=0.9\linewidth]{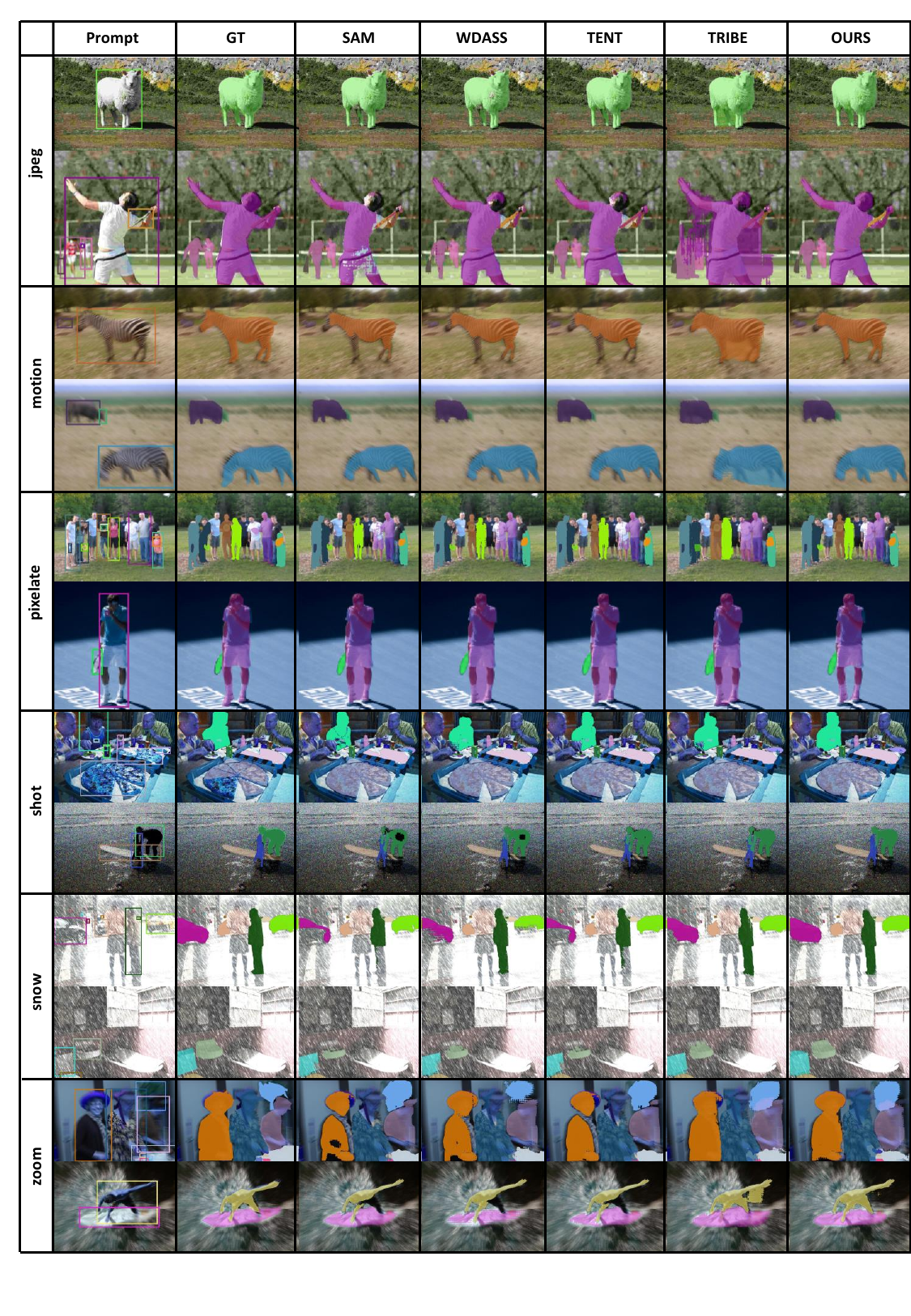}
    \vspace{-1cm}
    \caption{Visualization examples on COCO-C dataset.}
    \label{fig:corrupt-2}
\end{figure*}

Finally, we present more qualitative results on multiple downstream segmentation datasets. Specifically, COCO illustrations are shown in Fig.~\ref{fig:vis_coco}, and ones of ISIC are in Fig.~\ref{fig:vis_isic}, and ones of OCID are in Fig.~\ref{fig:vis_ocid}, and ones of CAMO are in Fig.~\ref{fig:vis_camo}, and ones of COCO-C are in Fig.~\ref{fig:corrupt-1},\ref{fig:corrupt-2}. The observations suggest SAM after weakly supervised adaptation achieves much superior segmentation quality on all types of weak supervision and testing prompts.

\end{document}